# Enhancing Kurdish Text-to-Speech with Native Corpus Training: A High-Quality WaveGlow Vocoder Approach

Abdulhady Abas Abdullah
Artificial Intelligence and Innovation Centre,
University of Kurdistan Hewlêr
Erbil, Kurdistan Region, Iraq

Sabat Salih Muhamad
Computer Science Department, Faculty of
Science, Soran University
Soran, Kurdistan, Iraq

Hadi Veisi
School of Intelligent Systems
College of Interdisciplinary Science and
Technologies University of Tehran, Iran

**Abstract**

The ability to synthesize spoken language from text has greatly facilitated access to digital content with the advances in text-to-speech technology. However, effective TTS development for low-resource languages, such as Central Kurdish (CKB), still faces many challenges due mainly to the lack of linguistic information and dedicated resources. In this paper, we improve the Kurdish TTS system based on Tacotron by training the Kurdish WaveGlow vocoder on a 21-hour central Kurdish speech corpus instead of using a pre-trained English vocoder WaveGlow. Vocoder training on the target language corpus is required to accurately and fluently adapt phonetic and prosodic changes in Kurdish language. The effectiveness of these enhancements is that our model is significantly better than the baseline system with English pretrained models. In particular, our adaptive WaveGlow model achieves an impressive MOS of 4.91, which sets a new benchmark for Kurdish speech synthesis. On one hand, this study empowers the advanced features of the TTS system for Central Kurdish, and on the other hand, it opens the doors for other dialects in Kurdish and other related languages to further develop.

## 1. Introduction

Text-to-speech has developed and evolved in the recent past, such that written text is converted to spoken language with most features of the intonation and cadence of a natural speaker (Tan et al., 2024); it is, therefore, an important application in many fields, for example, assistive technology for people with reading difficulties, interactive chatbots, advanced conversational AI, language learning platforms, or entertainment. At the heart of TTS research is speech synthesis, which is the ability to generate audio from text in a way that makes it natural speech for human level quality. Recent deep learning breakthroughs led to new models that would cover end-to-end speech synthesis. Speech synthesis technologies such as Char2Wav by Sotelo et al., 2017, Tacotron by Wang et al., 2017, Tacotron2 by Shen et al., 2018, DeepVoice by Arik et al., 2017, Transformer TTS by Li et al., 2019, FastSpeech by Ren et al., 2019, ParaNet by Peng et al., 2020, Neural Codec Language Models (NCLMs) by Wang et al. (2023), Matcha-TTS (Mehta et al., 2024). AI-Powered TTS engines are advancements that have uplifted synthesized speech quality beyond traditional statistical parametric methods. These recent works utilize the state-of-the-art neural vocoders, a source-filter vocoder like WORLD (Arthur and Csapó 2024), which can synthesize much better audio quality that sounds very similar to natural human speech.

However, despite these major advances in TTS, it is still a challenge for some languages to build high-quality and human-level systems, especially for low-resource languages, one of which is Kurdish (Abdullah et al., 2024). Kurdish, spoken by over 30 million people, is one such language. It is divided into three main dialects: Central Kurdish (CKB, i.e. Sorani), Northern Kurdish (Kurmanji), and Southern Kurdish (Abdullah and Veisi., 2022) (Veisi et al., 2024). In the last few years, several works have been done for the Kurdish language. In this direction, Muhamad and Veisi (2022) have worked on overcoming the problems of TTS in Kurdish by developing a

Central Kurdish corpus and using Tacotron 2 with the HiFi-GAN vocoder for synthesizing high-quality human-like speech. Using English character embeddings and 10 hours of recorded audio, the system was rated as having a mean opinion score of 4.1, which is similar to other state-of-the-art synthesizers in other languages. The more advanced end-to-end TTS model for the Kurdish language, in this case, for the Sorani dialect, introduced the use of variational auto-encoders combined with adversarial learning and a stochastic duration predictor. The latter enhances text-to-speech conversion by aligning the latent distributions and including diverse rhythms, hence yielding a high MOS of 3.94, which has outperformed all existing one-stage and two-stage systems with the measures used in naturalness and audio quality for RNN-TTS systems (Ahmad, H.A. and Rashid, T.A., 2024).

Another research paper proposes an end-to-end TTS system for Central Kurdish (CKB), hence handling data sparsity. Three different training experiments were performed for Tacotron2 on a dataset of 21 hours of female voice. The one trained from scratch on the full dataset scored the highest in terms of naturalness and intelligibility according to the Mean Opinion Score (MOS) of 4.78 out of 5 compared to other models (Muhamad et al., 2024). In addition, the development of the Kurdish TTS for the Central Kurdish dialect still needs to be improved. It is characterized as a low-resource language with different phonetic and linguistic characteristics, which Vocoder's general models do not adequately cover the characteristics of Kurdish speech due to the phonetic differences between the source language and the target language. The lack of a vocoder trained specifically for the Kurdish language is one of the obstacles so far to the Kurdish TTS model. In general, the relevance of designing dedicated vocoders that better represent Kurdish speech lies within the need for targeted development for improved quality in TTS of low-resource languages.

As a result., this paper makes several groundbreaking contributions to Kurdish speech synthesis. Firstly, it introduces the first TTS vocoder based on 21 hours of detailed speech data, which marks a significant advancement in Kurdish language technology. Secondly, we have successfully adapted the WaveGlow deep learning architecture (Prenger et al., 2019) to Kurdish, optimizing it for the unique acoustic properties of the language to ensure clear, natural speech output. Additionally, we have implemented advanced prosody modeling techniques to improve the rhythm, stress, and intonation of the synthesized speech, crucial for achieving lifelike speech quality. These enhancements not only push the boundaries of TTS for Kurdish but also offer scalable methodologies that can be applied to other Kurdish delict languages, broadening the impact of this work across different linguistic communities.

In the remainder of this paper, we elaborate on these themes in several sections: Section 2 discusses some recent advancements in TTS technology, focusing on Normalizing Flows and WaveGlow technology applied for developing low resource languages such as Kurdish. Section 3 presents the specific architecture of the adapted WaveGlow model for the Kurdish TTS. Section 4 presents the experimental results, and Section 5 concludes with a summary of our findings and some future lines that this area of research might take.

## 2. Related Works

With the advancements in neural network-based text-to-speech (TTS) systems, generating high-quality, natural-sounding speech has become increasingly feasible. However, because to a lack of data and resources, expanding TTS systems for low-resource languages like Kurdish continues to be difficult. This section examines relevant TTS research, emphasizing methods that make use of vocoder and its variants. We examine several approaches and works in the field, highlighting their applicability and possible modifications for Kurdish text-to-speech. Furthermore, we pinpoint the shortcomings and difficulties in the current literature, which directs our strategy for enhancing voice synthesis vocoders for low-resource languages.

Prenger et al. (2019) presented WaveGlow, a flow-based generative network for voice synthesis that integrates knowledge from WaveNet and Glow. With this approach the Mel spectrograms are transformed into high-quality speech using a single network and cost function. The realizations of WaveGlow are the efficient MOSs, the high sample rates of audio samples, and the equalized and consecutive training model. However, the model is very large and computationally expensive, which can be considered as major drawbacks of the presented approach.

Cui et al. h proposed a new glottal neural vocoder in 2019. It employs glottal-source filters combined with vocal-tract filters by utilizing the hybrid neural network. While this vocoder delivers good amount of voice synthesis that is of good quality, it comes alongside with implementation issues. Other research that can be mentioned in this context is the work of Csapó et al. (2019) that explored WaveGlow voice synthesis for ultrasound-based articulatory-to-acoustic mapping. WaveGlow, specifically developed for speakers and silent speech interfaces targeting clients with speaking disorders, outperforms regular vocoders and has more realistic synthesized speech, research showed. A major limitation is the use of ultrasonic technology.

Neural vocoder based on WaveNet was designed and presented by Oura et al. (2019); which is a very efficient neural vocoder attained especially for smooth real-time speech synthesizing. While this vocoder offers a very good quality synthesis it is very demanding in terms of the processing power needed. Extending the source-filter model into a flow-based deep generative model, which was named as ExcitGlow that is a variant of WaveGlow and it specifically targets on the distribution of the excitation signal rather than the speech signal. This approach contributes to a drastic reduction of the model size from 87. 73 million to 15. When choosing between GAN-based and TTS systems, our study determines that the latter can achieve an adequate model size of 60 million parameters while maintaining the perceivable quality of synthesized speech. Nevertheless, the closed-loop training framework that has been proposed by ExcitGlow is not easy to be adopted. The study completed by Al-Radhi et al. (2019) aims to demonstrated the application of RNNs to sequence-to-sequence for voice synthesis through utilizing a continuous vocoder. The findings of the study show that RNN-based models including LSTM, BLSTM, and GRU are more superior to feed-forward DNNs in aspects of naiveness and prediction capability and has the potential of producing the highest performance in SPSS. An inherent disadvantage in this method is the high computational complexities of RNN-based models which may cause slow loading. Some studies that have been made include Debnath et al.'s (2020) study used Tacotron2 and WaveGlow to assess the effectiveness of TTS systems involving Sanskrit in low-resource contexts. Their strategy was focused on transfer learning and fine-tuning of the available models to achieve near excellent voice synthesis with a limited amount of available data. The major limitation of this method is in its reliance on pre-trained models.

Ping and colleagues introduced WaveFlow for raw audio in 2020 in a flowing form that is concise and called flowing. This model adopts a dilated 2-D convolutional structure which captures long range structure as well as autoregressive functions in order to handle local changes. WaveFlow is significantly smaller and computational faster than WaveGlow that only requires 5. seventy-one billion ninety-one million parameters, and is capable of generating high-fidelity audio at a rate faster than real-time without the presence of a deliberately engineered inference kernel. However, the testing of this method has been limited to specific datasets, thus its application might be limited in other applications. Song et al. (2020) proposed efficient version of the WaveGlow, called Efficient WaveGlow or EWG for short This improvement is built on top of the previously mentioned model with primary intention of reducing number of parameters and increasing speed, while keeping the overall quality of the generated speech at descent level. In the EWG model, the WaveNet style transform network is replaced with the dilated convolution network utilizing FFTNet protocol. It also involves group convolution and it conveys local information between two layers. While it is conceivable that there may be trade-offs in quality for the large parameter reduction made, the improvements made are evident with alleviation in computing and inference time all while retaining high quality synthesis. The contribution of this work is the development of an enhanced technique for TTS synthesis for LRLs and provides a roadmap for further investigation.

A paper that is similar to the earlier study was done by Muhamad and Veisi in April of 2022 and plans on creating a Central Kurdish text-to-voice (TTS) system. The present system used transfer learning techniques borrowed from an English model that had been pre-trained. In addition, 10 hours of voice data in Central Kurdish were also used in total, and the HiFi-GAN vocoder was used. Many vocoders such as Griffin-Lim, WaveGlow, and HiFi-GAN are explored by Kumar et al., (2023) in order to examine their applied use in TTS systems that were designed for extensively spoken low resource languages with a focus on the Indian region. WaveGlow has been identified as famous for being capable of producing vocal output of superior quality, however, the authors explained that there are more enhanced vocoders notably; HiFi-GAN who are becoming popular due to efficiency and higher quality.

Ahmad and Rashid came up with new Terminated Text Synthesis (TTS) in Central Kurdish in 2024 based on end-to-end transformer. The suggested technique enhances the realism and intelligibility of the mapping of text-to-speech by learning mapping directly while using data augmentation methods to solve a problem of data shortage. Unquestionably, such an approach was quite useful; nonetheless, the method demanded much processing capacity and its success depended on the quality of the available corpus. In another previous study (Muhamad et al., 2024), we employed transfer learning from an English pre-trained model to train our initial model. However, we utilized a larger dataset (21 h) and substituted the HiFi-GAN vocoder with the WaveGlow vocoder, which was trained on an English pre-trained model. This modification resulted in the generation of speech with superior quality and a more authentic sound. The summary of the related studies is provided, highlighting their important elements in Table 1.

Table 1: Summary of review on Vocoder technique for Speech Synthesis.

| Reference | Methodology | Advantages | Disadvantages |
|---|---|---|---|
| **Cui et al. 2018** | Glottal source + vocal tract filters, hybrid neural network | High-quality, efficient synthesis | Implementation complexity |
| **Prenger et al., 2019** | Glow + WaveNet, single cost function | High-quality speech, stable training | High computational cost, large model size |
| **Al-Radhi et al., 2019** | Sequence-to-sequence RNNs | Improved naturalness, state-of-the-art performance | Computationally intensive |
| **Oura et al. 2019** | WaveNet-style transform, optimized vocoder | Real-time, high-quality synthesis | High resource requirements |
| **Song et al., 2020** | FFTNet-style, group convolution, shared local conditions | Reduced computation, faster inference, smaller model | Possible quality trade-offs |
| **Ping et al., 2020** | 2-D convolutions, autoregressive functions | Smaller model, faster synthesis, high fidelity | Limited dataset testing |
| **Oh et al., 2020** | Source-filter model, flow-based generative model | Reduced model size, high-quality synthesis | Complex training framework |
| **Debnath et al. 2020** | Transfer learning, fine-tuning Tacotron2 + WaveGlow | High-quality synthesis with limited data | Dependency on pre-trained models |
| **Csapó et al. 2020** | Ultrasound tongue images to Mel-spectrograms | Improved naturalness for silent speech interfaces | Requires ultrasound equipment |
| **Muhamad and Veisi, 2022** | Transfer learning, Tacotron2 + HiFi-GAN | High-quality synthesis with limited data | Dependency on pre-trained models |
| **Kumar et al. (2023)** | Tacotron 2 + WaveGlow, Griffin-Lim, and HiFi-GAN vocoders | Applied several vocoders, conducting a comparative analysis. | Data quality obtained from crowdsourcing can exhibit variability. |

| | | Protocols for collecting scalable data. | Restricted discourse on difficulties particular to language. Vocoders of superior quality may experience scaling challenge. |
|---|---|---|---|
| **Ahmad, H.A. and Rashid, T.A. (2024)** | end-to-end Transformer model | High-quality synthesis Better Performance in Low-Resource Settings | High Computational Requirements Dependency on Data Quality |
| **Muhamad et al., 2024** | Tacotron2 + WaveGlow | High-quality speech | Dependency on pre-trained models |

In conclusion, for low-resource languages like Kurdish, selecting an appropriate vocoder is crucial to balance quality and efficiency. Based on a review of previous works, WaveGlow stands out as a suitable choice. WaveGlow significantly reduces computational costs and model size while maintaining high-quality speech synthesis, making it well-suited for resource-constrained environments.

## 3. Methodology

In this section we discuss the speech corpus for the target language, with the proposed approach for enhancing TTS by training new WaveGlow for Kurdish language

### 3.1 Kurdish Speech Corpus

**Train phase:** In this study, we utilized the existing "Sabat Speech Corpus" (Muhamad et al., 2024) for training the vocoder WaveGlow, enhancing its capability to synthesize Central Kurdish speech. The corpus, which was previously compiled and contains 10,979 utterances across diverse categories such as news, sports, linguistics, psychology, poetry, health, scientific topics, general knowledge, interviews, politics, education, literature, narratives, tourism, and miscellaneous subjects.

Table 2 presents a comprehensive breakdown of the different types of utterances and their respective counts, demonstrating the wide range of topics covered in our corpus.

Table 2: The total of the train utterances (Muhamad et al., 2024)

| Category | No. of Utterances |
|---|---|
| **linguistics** | 1760 |
| **questions and exclamation** | 1393 |
| **story** | 1092 |
| **poem** | 916 |
| **tourism** | 782 |
| **miscellaneous** | 700 |
| **sport** | 683 |
| **education and literature** | 619 |
| **news** | 608 |
| **science** | 543 |
| **health** | 483 |
| **politics** | 483 |
| **general information** | 461 |
| **interview** | 456 |
| **Total** | 10,979 |

This broad coverage is crucial for training a vocoder like WaveGlow, as it ensures that the system can accurately reproduce a wide range of phonetic and intonational nuances inherent to the Kurdish language. By leveraging this pre-existing corpus, we aim to improve the naturalness and expressiveness of synthesized Kurdish speech, addressing the unique challenges of speech synthesis in low-resource language settings. Summary Overview of the Sabat Speech Corpus is shown in Table 3.

Table 3: Overview of the Sabat Speech Corpus (Muhamad et al., 2024)

| Feature | Details |
|---|---|
| Total Utterances | 10,979 utterances |
| Audio Length | 21 hours total |
| Sampling Rate | 22,050 Hz |
| Bit Depth | 16 bits |
| Channel | Mono |
| File Format | WAV |
| Categories | News, Sport, Linguistics, Poem, Health, etc. |
| Recording Environment | Professional studio |
| Speaker Profile | Female, from Sulaymaniyah, in her thirties |
| Text Normalization | Applied AsoSoft Library Normalization tool[1] |

**Test Phase:** For the test set, 110 sentences were meticulously selected from texts covering 17 distinct subject areas (Muhamad et al., 2024), ensuring they were different from the training sentences to effectively gauge the TTS model's performance across diverse contexts. These sentences were sourced from several websites and refined to align with Central Kurdish orthography standards. This careful curation aims to test the TTS system's ability to handle a variety of linguistic challenges and assess its generalization capabilities across different types of content listed in Table 4.

Table 4: Distribution of Test Set Sentences Across Topics (Muhamad et al., 2024)

| Topics | No. of Sentences |
|---|---|
| News | 10 |
| Formal Letter | 10 |
| Sport | 9 |
| Poem | 8 |

---

[1] https://github.com/AsoSoft/AsoSoft-Library

| | |
|---|---|
| Questions | 7 |
| Psychology | 6 |
| Health | 6 |
| Science | 6 |
| Miscellaneous | 6 |
| General Information | 6 |
| Story | 6 |
| Tourism | 6 |
| Linguistics | 5 |
| Interview | 5 |
| Politics | 5 |
| Education and Literature | 5 |
| Exclamation | 4 |
| **Total** | **110** |

## 3.2 Text to Speech

The general design for this TTS model involves a two-stage process (Li et al., 2024). The first model converts text data into Mel spectrograms, and the second model takes these Mel spectrograms as input, producing binary sound wave representations. These sound waves, when played, generate human speech. This structure is illustrated in the accompanying diagram and showed in Figure 1.

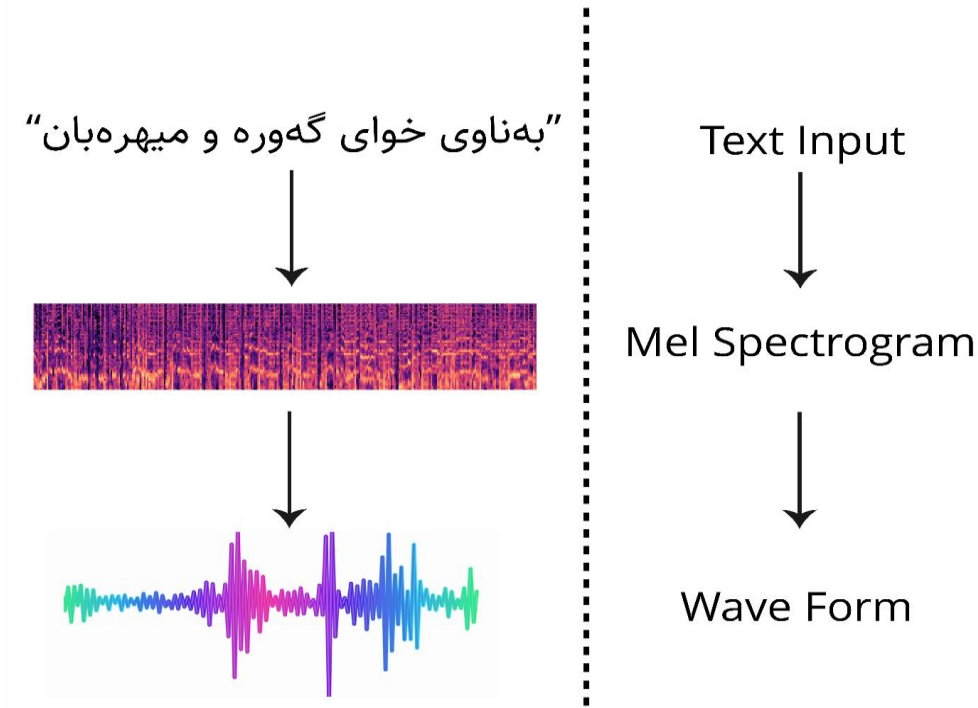

Figure 1: Text to Mel to Waveform

### 3.2.1 Tacotron 2

The methodology presented in our previous work (Muhamad et al., 2024), as shown in Figure 2, entails generating a dataset consisting of "text, audio" pairs from a single female speaker. This dataset is then subjected to preprocessing procedures, including text normalization, in order to prepare the data for our TTS system. For the prediction of Mel spectrograms, we utilized a sequence-to-sequence synthesis network that is based on Tacotron2 (Shen et al., 2018a). These spectrograms are then transformed into audible sounds using the WaveGlow vocoder (Prenger et al., 2019). Tacotron2's streamlined end-to-end design, including its acoustic model and vocoder stages, is well-suited for Kurdish because of its straightforward phonetic structure, which requires less input compared to more intricate languages. The proposed model utilized a recurrent neural network to produce Mel spectrograms from given textual input. Subsequently, the WaveGlow vocoder converts these spectrograms into waveforms. The Tacotron2 model and WaveGlow vocoder are trained separately on our comprehensive Kurdish voice corpus.

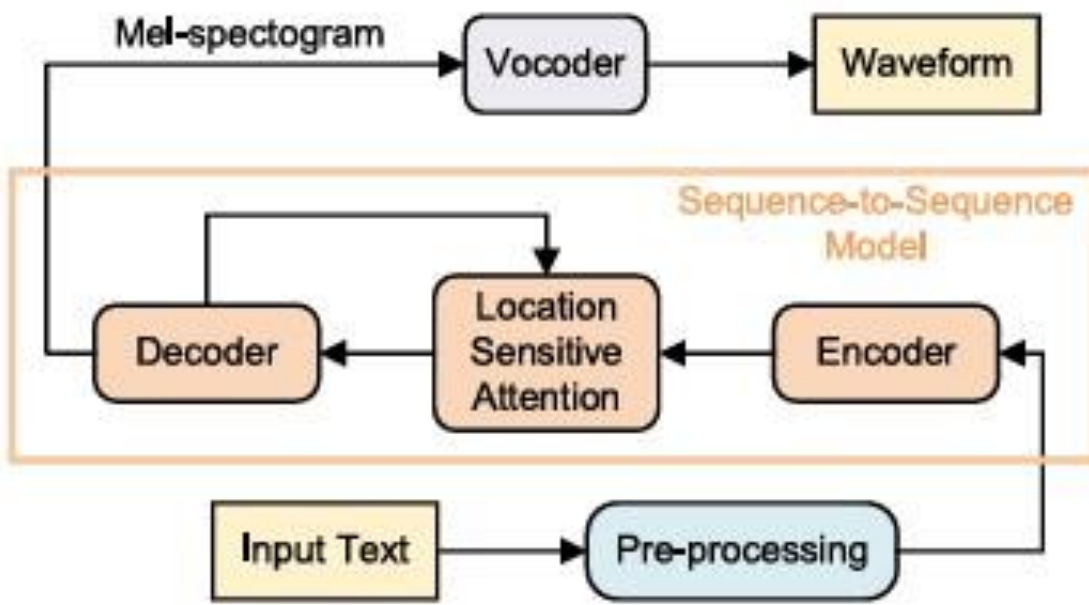

Figure 2: The block diagram from our previously published methodology (Muhamad et al., 2024)

### 3.2.2 Vocoder Architecture

Figure 3 shows the conversion of the Mel spectrogram into speech waveform with help of WaveGlow vocoder. The first input is a Mel spectrogram, which graphically depicts the density of the frequencies in audio signal with respect to the time. As it will be explained later in this paper, the Mel spectrogram retains features of the audio that are critical for signal processing and analysis because they correspond with characteristics of human auditory processing. Subsequently, the Mel spectrogram undergoes the process called "Vocoder WaveGlow" which forms the end part of the proposed model. WaveGlow is a state-of-art generative model, aimed at the high-quality audio generation from Mel spectrograms (Debnath et al., 2020). A set of invertible transformations, the normalizing flows, is used to transform the data, the Mel spectrogram that has a simpler distribution than the distribution of the audio waveform. In so doing, WaveGlow transforms the Mel spectrogram back into the original signal and all its details into a waveform. The final product from the WaveGlow vocoder is a "Speech Waveform". This waveform is the reconstructed audio signal which can be played and heard by human intervention as speeches being made. It has all the differential attributes of natural speech that is synthesized from the features embedded in Mel spectrogram. The speech waveform is the result that listeners hear, and it is generated to closely mimic natural human speech. It depicts the progression from the Mel spectrogram, which is a text-based or visual representation of sound frequencies, to the speech waveform, which is the audible output. The below figure demonstrates how the WaveGlow model bridges the gap between the abstract representation of audio in the form of a Mel spectrogram and the concrete, listenable speech waveform. The process underscores the capability of the WaveGlow vocoder to synthesize high-quality, natural-sounding speech from a structured, frequency-based input.

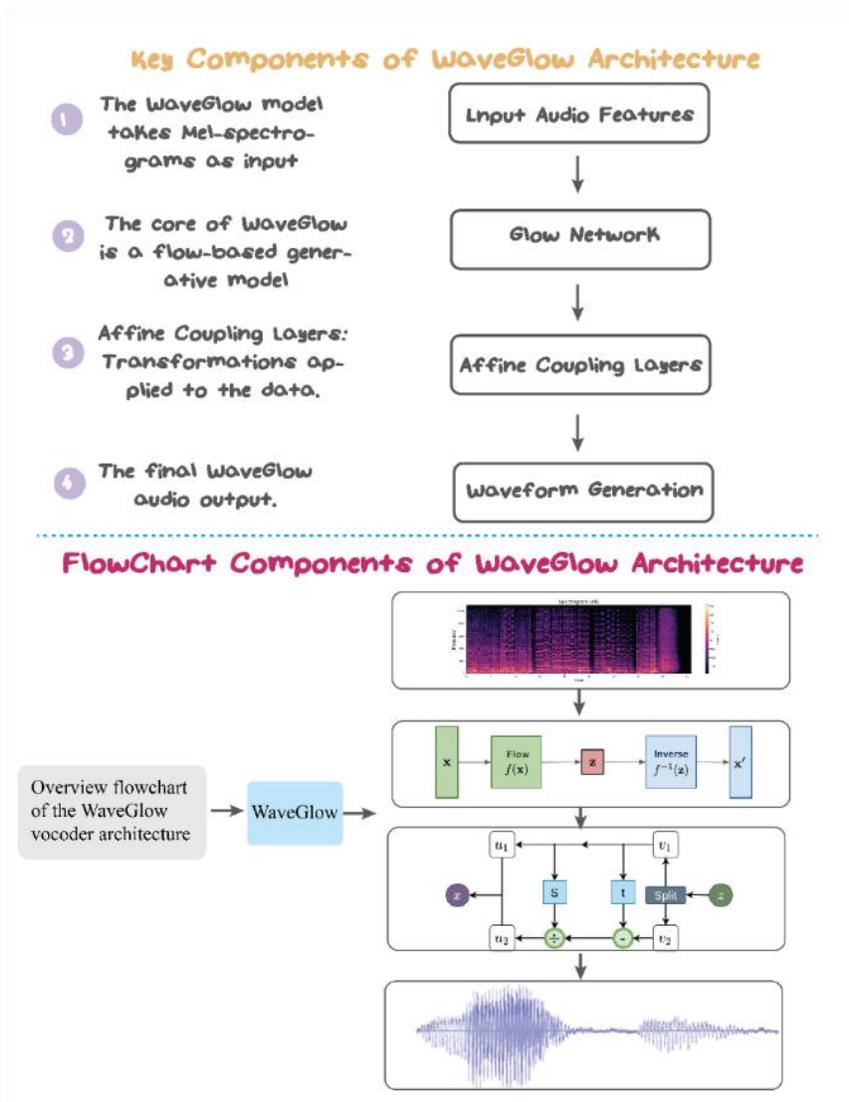

Figure 3: Overview key components and flowchart vocoder WavGlow Architecture

## A. WaveGlow

WaveGlow, inspired by the Glow concept, offers faster synthesis compared to WaveNet (Oord et al., 2016). It employs an invertible transformation between blocks of eight time-domain audio samples and a standard normal distribution, conditioned on the log Mel-spectrogram (Mustafa et al., 2021). This transformation enables audio generation by sampling from this Gaussian density. The invertible transformation consists of a sequence of individual invertible transformations, known as normalizing flows. In WaveGlow, each flow is composed of a 1x1 convolutional layer followed by an affine coupling layer. The affine coupling layer is a neural transformation that predicts a scale and bias based on the input speech $x$ and Mel-spectrogram $X$. Let $W\_k$ represent the learned weight matrix for the $k-th$ 1x1 convolutional layer and $s\_j(x, X)$ be the predicted scale value at the $j-th$ affine coupling layer. For inference, WaveGlow samples z from a uniform Gaussian distribution and applies the inverse transformations ($f^{-1}$) conditioned on the Mel spectrogram X to retrieve the speech sample x. Due to the simplicity of parallel sampling from the Gaussian distribution, all audio samples can be generated simultaneously. The model is trained to minimize the log-likelod of the clean speech samples x:

$$lnP(x|X) = lnP(z) - \sum j=0Jlns_j(x,X) - \sum k=0Kln|W_k| \qquad Eq(1)$$

where $J$ is the number of coupling transformations and $K$ is the number of convolution layers. In $P(z)$, the parameter $v^2$ refers to the variance of the Gaussian distribution, and $v = 1$ is used during training. WaveGlow uses an alternative to the logistic function in equation (1) for sampling. This system was implemented using 12 coupling layers, each with 8 layers of dilated convolution with 512 residual channels. The same architecture was adapted for the PR system using WaveGlow as its vocoder, referred to as PR-WaveGlow. The architecture of the WaveGlow model is shown in Figure 4.

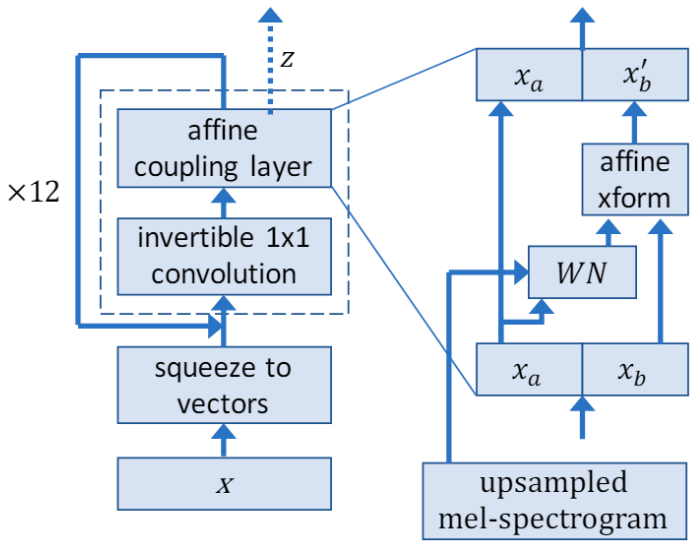

Figure 4: The architecture of the WaveGlow model (Prenger et al., 2019).

## 4 Results and Discussion

### 4.1 TTS measure

MOS is a commonly used measure in evaluating the quality of synthesized speech (Maguer et al., 2024). It involves human listeners rating the quality of speech samples on a predefined scale, typically from 1 to 5, where:

- 1: Bad
- 2: Poor
- 3: Fair
- 4: Good

- 5: Excellent

The MOS is calculated by averaging the scores given by all listeners for each speech sample. Equation (2) is utilized to ascertain the assessment of intelligibility and naturalness for MOS.

$$\text{MOS} = \frac{1}{N}\Sigma \sum_{i=0}^{N} s_i \qquad Eq(2)$$

Where:

- $N$ is the number of listeners.
- $s_i$ is the score given by the $i-th$ listener.

For our evaluation, native Kurdish speakers rated the naturalness and intelligibility of the synthesized speech samples.

## 4.2 Implementation Tools and Hyper-Parameters

The training and implementation of the Kurdish WaveGlow model were conducted using two NVIDIA RTX 4090 GPUs, providing a total of 48 GB of GPU memory. The system also utilized a total of 290 GB of RAM. The training parameters used for developing the Kurdish WaveGlow model are summarized in Table 5.

Table 5: Training Parameters

| Parameter | Value |
| --- | --- |
| Batch Size | 22 |
| Learning Rate | 1e-4 with an exponential decay |
| Optimizer | Adam optimizer (beta1=0.9, beta2=0.999) |
| Number of Epochs | 100000 |
| Sigma | 1.0 |
| Iterations per Checkpoint | 2000 |
| Seed | 1234 |

The parameters in Table 5 describe the settings used to train the Kurdish WaveGlow model. The batch size indicates the number of samples processed before the model is updated. The learning rate defines the step size for weight updates. The optimizer used is Adam, known for its efficiency and adaptive learning rate capabilities. Training was conducted for 100,000 epochs to ensure comprehensive learning, with intermediate checkpoints saved every 2000 iterations. A fixed seed ensures the reproducibility of the results.

Table 6: Data Configuration Parameters

| Parameter | Value |
|---|---|
| **Segment Length** | 16000 |
| **Sampling Rate** | 22050 Hz |
| **Filter Length** | 1024 |
| **Hop Length** | 256 |
| **Win Length** | 1024 |
| **Mel Frequency Min** | 0.0 |
| **Mel Frequency Max** | 8000.0 |

Table 6 details the data configuration parameters used during the preprocessing stage. These settings ensure the audio data is correctly formatted and segmented for training. The segment length specifies the number of audio samples per segment, while the sampling rate defines the number of samples per second. Filter length, hop length, and win length are used in the Short-Time Fourier Transform (STFT) process to convert audio signals into Mel-spectrograms. Mel frequency parameters set the range of frequencies to be included in the Mel-spectrogram.

Table 7: WaveGlow Configuration Parameters

| Parameter | Value |
|---|---|
| Number of Mel Channels | 80 |
| Number of Flows | 12 |
| Group Size | 8 |
| Early Every | 4 |
| Early Size | 2 |
| - Layers | 8 |
| - Channels | 256 |
| - Kernel Size | 3 |

Table 7 provides the configuration parameters specific to the WaveGlow model. The number of Mel channels indicates the dimensionality of the Mel-spectrogram features. The number of flows specifies the layers of transformations used to map the Mel-spectrogram to audio. Group size determines the number of samples processed together in the model, while early every and early size parameters control the early output of samples to improve efficiency. The WaveNet configuration includes the number of layers, channels, and kernel size used in the WaveGlow network to generate high-fidelity audio.

**4.3 Experimental Results**

The Kurdish WaveGlow model was trained from scratch using the Sabat Speech Corpus, which consists of 21 hours of high-quality, annotated speech data. The training process did not involve any pre-trained models or fine-tuning; instead, the model was developed entirely from the ground up to ensure it could fully adapt to the unique characteristics of the Kurdish language. The training process spanned a total of 5 days, with each day comprising 24 hours of continuous training, amounting to 120 hours in total. The model was configured to save checkpoints every 2000 iterations, resulting in a total of 2,702,000 checkpoints throughout the training period. This extensive training and checkpointing allowed for meticulous tracking of the model's performance and adjustments as needed. Throughout the training period, the model demonstrated steady convergence, with a consistent decrease in training loss. This indicates that the model effectively learned the acoustic properties and linguistic nuances of Kurdish speech from the Sabat Corpus.

**4.4 Evaluations**

For the MOS evaluation, 110 random sentences from various categories were selected, ensuring these sentences were not part of the training set. The categories included news, sports, linguistics, psychology, poetry, health, questions, exclamations, science, miscellaneous, general information, interviews, politics, education and literature, stories, tourism, and SMS. This diverse selection ensured a comprehensive assessment of the model's performance across different types of content.

- **Raters**: Twelve native Kurdish speakers (7 males and 5 females, aged 21 to 46) participated in the MOS evaluation. They listened to the synthesized sentences using headphones to ensure consistent audio quality and rated each sentence on the 5-point Likert scale.

The results of the MOS evaluation for each category for the Kurdish Tacotron2-Scratch (Muhamad et al., 2024) WaveGlow Kurdish-Scratch model are shown in Figure 5.

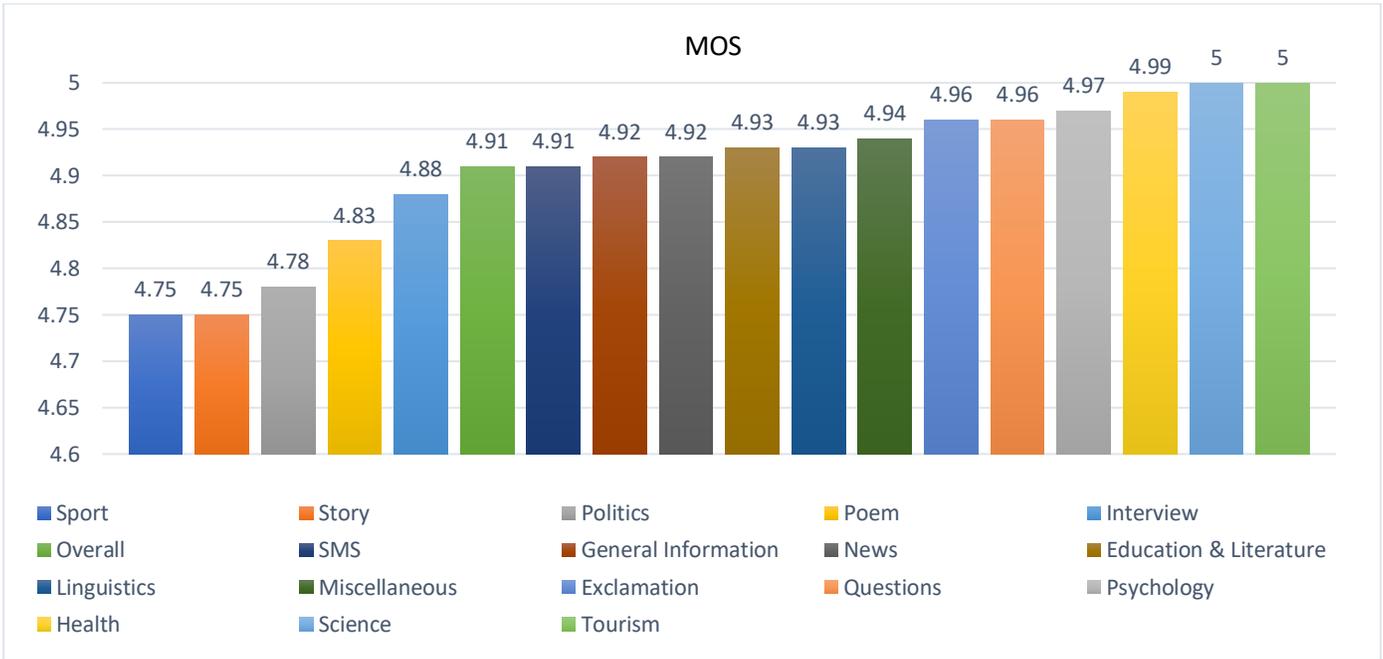

Figure 5: MOS Results for Kurdish Tacotron2-Scratch (Muhamad et al., 2024) WaveGlow Kurdish-Scratch

These MOS results highlight the high performance of the Kurdish Tacotron2-Scratch (Muhamad et al., 2024) WaveGlow Kurdish-Scratch model across various content categories, reflecting its ability to generate natural and intelligible Kurdish speech.

## 4.5 Comparative Analysis of Kurdish TTS: WaveGlow (Kurdish) vs. Vocoders (English Pre-trained)

This time the comparison was made between the TTS models trained from scratch on the Kurdish-specific corpus and the models incorporated with the English pre-trained WaveGlow. The comparison is made in view of the average MOS of all the models and also of each genre and sub-genre of contents. This analysis gives performance information of the synthesized speech and the extent of naturalness of the synthesized speech by the different training approaches. Table 8 below also demonstrates the MOS for different content classifications. These results highlight the performance differences between genuine voice samples and the three models: Kurdish Tacotron2 pre-trained with HiFi-GAN English pre-trained and Kurdish Tacotron2 scratch with WaveGlow English pre-trained and the proposed model Kurdish Tacotron2 scratch with WaveGlow vocoder that train for Kurdish dataset only.

Table 8: MOS Results for Different Kurdish Models Benchmark by Category

| Category | MOS (Genuine Voice) | Kurdish Tacotron2-Pretrain with HiFi-GAN English Pre-trained (Muhamad, and Veisi., 2022) | Kurdish Tacotron2-Scratch with WaveGlow English Pre-trained (Muhamad et al., 2024) | Our model (Kurdish Tacotron2-Scratch with WaveGlow Kurdish-Scratch) |
|---|---|---|---|---|
| News | 5.0 | 4.2 | 4.5 | 4.92 |
| Sports | 4.9 | 4.1 | 4.4 | 4.75 |
| Linguistics | 4.8 | 4.0 | 4.3 | 4.93 |
| Psychology | 5.0 | 4.3 | 4.6 | 4.97 |
| Poem | 4.9 | 4.2 | 4.5 | 4.83 |
| Health | 4.8 | 4.1 | 4.4 | 4.99 |
| Questions | 5.0 | 4.2 | 4.5 | 4.96 |
| Exclamation | 5.0 | 4.2 | 4.5 | 4.96 |
| Science | 4.9 | 4.1 | 4.4 | 5.00 |
| Miscellaneous | 4.8 | 4.0 | 4.3 | 4.94 |
| General Info | 4.9 | 4.2 | 4.5 | 4.92 |
| Interviews | 4.9 | 4.2 | 4.5 | 4.88 |
| Politics | 5.0 | 4.3 | 4.6 | 4.78 |
| Education & Lit | 4.8 | 4.0 | 4.3 | 4.93 |
| Story | 4.9 | 4.2 | 4.5 | 4.75 |
| Tourism | 4.9 | 4.2 | 4.5 | 5.00 |
| SMS | 4.8 | 4.0 | 4.3 | 4.91 |

These results demonstrate that the Tacotron2-Scratch (WaveGlow Kurdish-Scratch) model consistently outperformed the other models across all categories. The genuine voice samples, as expected, received the highest MOS scores. However, the Kurdish-Scratch model closely followed, indicating a high level of naturalness and intelligibility in the synthesized speech. To provide a comprehensive comparison, the overall MOS results for each model were averaged across all categories. Figure 6 analysis highlights the general performance and quality of the synthesized speech for each model.

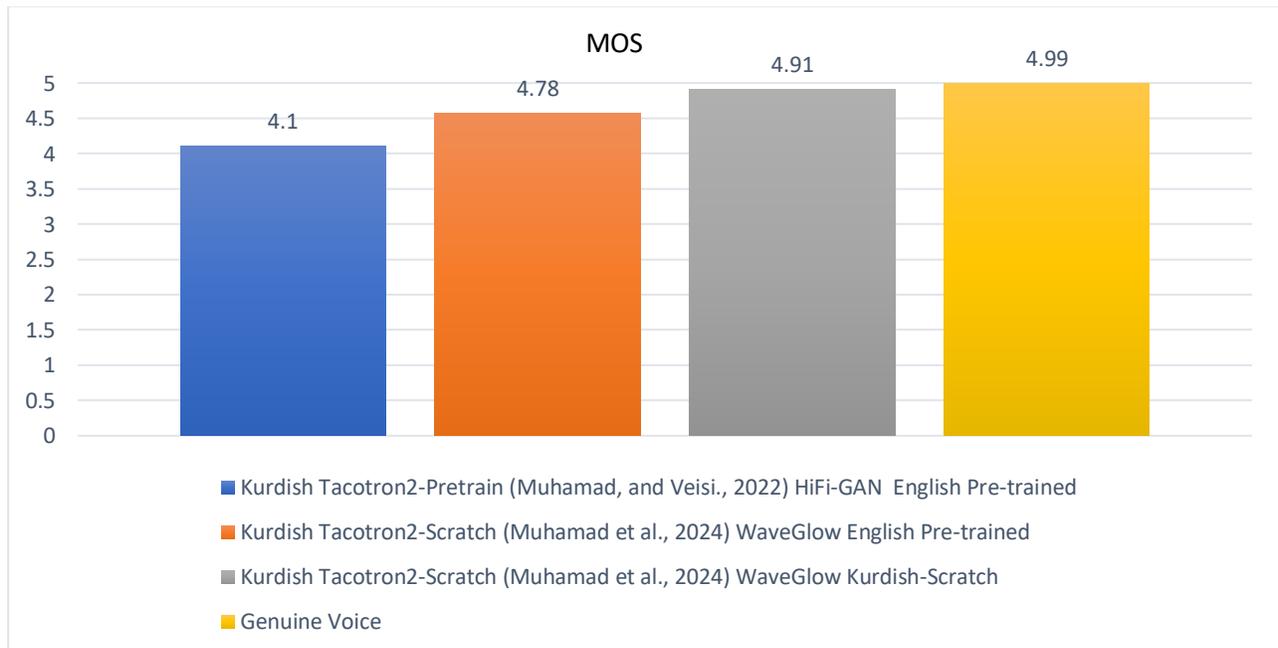

Figure 6: Comparative Analysis of MOS Results for Kurdish Models Benchmark

The MOS results clearly indicate that the Tacotron2-Scratch (WaveGlow Kurdish-Scratch) model achieved significantly higher scores compared to the models based on English pre-trained WaveGlow. The Kurdish-Scratch model's MOS of 4.78 highlights its ability to generate highly natural and intelligible Kurdish speech, closely approximating genuine voice samples. The lower MOS scores for the Tacotron2-Pre-train (HiFi-GAN) English Pre-trained (Muhamad, and Veisi., 2022) model reflect the limitations of adapting a pre-trained model to a low-resource language like Kurdish. While it still performed reasonably well, it struggled with pronunciation and naturalness compared to the Kurdish-specific model trained from scratch. Overall, these results emphasize the importance of training TTS models from scratch using language-specific data to achieve the highest quality of synthesized speech. The Kurdish-Scratch model's superior performance demonstrates the effectiveness of this approach for low-resource languages.

## 5. Summery and Future Works

This work has made significant improvements for TTS technology for Kurdish language especially for Kurdish WaveGlow vocoder to synthesis Kurdish speech. The final outcome of our study is the integration of an individual TTS system with the help of 21hrs of clean labelled speech from the Sabat Speech Corpus. This system is therefore devised to accommodate the architectural and development feature of the Kurdish language which is both a popular teaching language and a restricted resource language. These improvements were made through the application of more advanced techniques of prosody modeling to the WaveGlow model: experimental modifications to synthesize Kurdish. All these improvements are thus necessary in order to achieve an accurate and realistic quality. Besides enhancing the intelligibility and the naturalness of the voice output, the obtained TTS system generates a versatile paradigm that can be extended to other LR languages and, consequently, may revolutionize TTS solutions in overlooked linguistic environments. They also statistically substantiate that our Kurdish-specific proposed model has relatively larger (MOS) in various contents compared to the models with employ the English pre-trained systems identified in the experiments. This underlines the inherent problem of the lack of language-specific TTS models as it is the only way to ensure the correctness and accurate translation of generated speech. The successful implementation and evaluation of this Kurdish TTS system set the benchmark for future works inside the field of speech synthesis and particularly for languages with limited resources. This precedes possibilities to create speech-based systems that are more accessible to those with disabilities, as well as caters for the fact that people across the globe use different languages. Thus, in the future, the focus will be on improving these approaches, expanding the data base including various linguistic data and increasing the efficiency of the system to relate to the complex linguistic and acoustic challenges of the Kurdish language and other low-resource languages. Besides extending the progress of technology in the enhancement of Kurdish digital resources our project also contributes to the discourse on language conservation and digital opportunity.

## References

Abdullah, A.A. and Veisi, H., 2022. Central Kurdish Automatic Speech Recognition using Deep Learning. Journal of University of Anbar for Pure Science, 16(2).

Abdullah, A.A., Veisi, H. and Rashid, T., 2024. Breaking Walls: Pioneering Automatic Speech Recognition for Central Kurdish: End-to-End Transformer Paradigm. arXiv preprint arXiv:2406.02561.

Ahmad, H.A. and Rashid, T.A., 2024. Central Kurdish Text-to-Speech Synthesis with Novel End-to-End Transformer Training. *Algorithms*, *17*(7), p.292.

Arik, S. Ö., Chrzanowski, M., Coates, A., Diamos, G., Gibiansky, A., Kang, Y., ... & Shoeybi, M. (2017). Deep Voice: Real-time Neural Text-to-Speech. In *Proceedings of ICML 2017*.

Arthur, F.V. and Csapó, T.G., 2024. Speech synthesis from intracranial stereotactic Electroencephalography using a neural vocoder. *INFOCOMMUNICATIONS JOURNAL: A PUBLICATION OF THE SCIENTIFIC ASSOCIATION FOR INFOCOMMUNICATIONS (HTE)*, *16*(1), pp.47-55.

Csapó, T.G., Zainkó, C., Tóth, L., Gosztolya, G. and Markó, A., 2020. Ultrasound-based articulatory-to-acoustic mapping with WaveGlow speech synthesis. arXiv preprint arXiv:2008.03152.

Cui, Y., Wang, X., He, L. and Soong, F.K., 2018, September. A New Glottal Neural Vocoder for Speech Synthesis. In *Interspeech* (pp. 2017-2021).

Debnath, A., Patil, S.S., Nadiger, G. and Ganesan, R.A., 2020, December. Low-resource end-to-end sanskrit tts using tacotron2, waveglow and transfer learning. In *2020 IEEE 17th India Council International Conference (INDICON)* (pp. 1-5). IEEE.

Kumar, G.K., Praveen, S.V., Kumar, P., Khapra, M.M. and Nandakumar, K., 2023, June. Towards building text-to-speech systems for the next billion users. In *ICASSP 2023-2023 IEEE International Conference on Acoustics, Speech and Signal Processing (ICASSP)* (pp. 1-5). IEEE.

Le Maguer, S., King, S. and Harte, N., 2024. The limits of the Mean Opinion Score for speech synthesis evaluation. *Computer Speech & Language*, *84*, p.101577.

Li, N., Liu, S., Liu, Y., Zhao, S., & Liu, M. (2019). Neural Speech Synthesis with Transformer Network. In *Proceedings of AAAI 2019*.

Li, Y.A., Han, C., Raghavan, V., Mischler, G. and Mesgarani, N., 2024. Styletts 2: Towards human-level text-to-speech through style diffusion and adversarial training with large speech language models. *Advances in Neural Information Processing Systems*, *36*.

Mehta, S., Tu, R., Beskow, J., Székely, É. and Henter, G.E., 2024, April. Matcha-TTS: A fast TTS architecture with conditional flow matching. In *ICASSP 2024-2024 IEEE International Conference on Acoustics, Speech and Signal Processing (ICASSP)* (pp. 11341-11345). IEEE.

Morise, M., Yokomori, F. and Ozawa, K., 2016. World: a vocoder-based high-quality speech synthesis system for real-time applications. *IEICE TRANSACTIONS on Information and Systems*, *99*(7), pp.1877-1884.

Muhamad, S. and Veisi, H., 2022. End-to-End Kurdish Speech Synthesis Based on Transfer Learning. *Passer Journal of Basic and Applied Sciences*, *4*(2), pp.150-160.

Muhamad, S.S., Veisi, H., Mahmudi, A., Abdullah, A.A. and Rahimi, F., 2024. Kurdish end-to-end speech synthesis using deep neural networks. *Natural Language Processing Journal*, p.100096.

Mustafa, A., Pia, N. and Fuchs, G., 2021, June. Stylemelgan: An efficient high-fidelity adversarial vocoder with temporal adaptive normalization. In *ICASSP 2021-2021 IEEE International Conference on Acoustics, Speech and Signal Processing (ICASSP)* (pp. 6034-6038). IEEE.

Oh, S., Lim, H., Byun, K., Hwang, M.J., Song, E. and Kang, H.G., 2020, December. ExcitGlow: Improving a WaveGlow-based neural vocoder with linear prediction analysis. In *2020 Asia-Pacific Signal and Information Processing Association Annual Summit and Conference (APSIPA ASC)* (pp. 831-836). IEEE.

Oord, A.V.D., Dieleman, S., Zen, H., Simonyan, K., Vinyals, O., Graves, A., Kalchbrenner, N., Senior, A. and Kavukcuoglu, K., 2016. Wavenet: A generative model for raw audio. *arXiv preprint arXiv:1609.03499*.

Oura, K., Nakamura, K., Hashimoto, K., Nankaku, Y. and Tokuda, K., 2019. Deep neural network-based real-time speech vocoder with periodic and aperiodic inputs. *Proc. SSW10*, *32*, pp.13-18.

Peng, K., Ping, W., Song, Z., & Zhao, K. (2020). Non-Autoregressive Neural Text-to-Speech. In *Proceedings of ICML 2020*.

Ping, W., Peng, K., Zhao, K. and Song, Z., 2020, November. Waveflow: A compact flow-based model for raw audio. In *International Conference on Machine Learning* (pp. 7706-7716). PMLR.

Prenger, R., Valle, R. and Catanzaro, B., 2019, May. Waveglow: A flow-based generative network for speech synthesis. In ICASSP 2019-2019 IEEE International Conference on Acoustics, Speech and Signal Processing (ICASSP) (pp. 3617-3621). IEEE.

Ren, Y., Ruan, Y., Tan, X., Qin, T., Zhao, S., & Liu, T. (2019). FastSpeech: Fast, Robust, and Controllable Text to Speech. In *Proceedings of NeurIPS 2019*.

Salah Al-Radhi, M., Gábor Csapó, T. and Németh, G., 2019. RNN-based speech synthesis using a continuous sinusoidal model. *arXiv e-prints*, pp.arXiv-1904.

Shen, J., Pang, R., Weiss, R. J., Schuster, M., Jaitly, N., Yang, Z., ... & Zen, H. (2018). Natural TTS Synthesis by Conditioning Wavenet on Mel Spectrogram Predictions. In *Proceedings of ICASSP 2018*.

Song, W., Xu, G., Zhang, Z., Zhang, C., He, X. and Zhou, B., 2020. Efficient WaveGlow: An Improved WaveGlow Vocoder with Enhanced Speed. In *INTERSPEECH* (pp. 225-229).

Sotelo, J., Mehri, S., Kumar, K., Santos, J. F., Kastner, K., Courville, A., & Bengio, Y. (2017). Char2Wav: End-to-End Speech Synthesis. In *Proceedings of ICLR 2017*.

Tan, X., Chen, J., Liu, H., Cong, J., Zhang, C., Liu, Y., Wang, X., Leng, Y., Yi, Y., He, L. and Zhao, S., 2024. Naturalspeech: End-to-end text-to-speech synthesis with human-level quality. *IEEE Transactions on Pattern Analysis and Machine Intelligence*.

Veisi, H., muhealddin Awlla, K. and Abdullah, A.A., 2024. KuBERT: Central Kurdish BERT Model and Its Application for Sentiment Analysis.

Wang, C., Chen, S., Wu, Y., Zhang, Z., Zhou, L., Liu, S., Chen, Z., Liu, Y., Wang, H., Li, J. and He, L., 2023. Neural codec language models are zero-shot text to speech synthesizers. *arXiv preprint arXiv:2301.02111*.

Wang, Y., Skerry-Ryan, R., Stanton, D., Wu, Y., Weiss, R. J., Jaitly, N., ... & Le, Q. V. (2017). Tacotron: Towards End-to-End Speech Synthesis. In *Proceedings of Interspeech 2017*.